\documentclass[runningheads]{llncs}
\usepackage[T1]{fontenc}
\usepackage{graphicx}
\usepackage{amsmath}
\usepackage{hyperref}
\usepackage{rotating}
\begin{document}
\title{What-If Explanations Over Time: Counterfactuals for Time Series Classification}
\titlerunning{What-If Explanations Over Time}
%
\author{Udo Schlegel\inst{1,2}\orcidID{0000-0002-8266-0162} \and Thomas Seidl\inst{1,2}\orcidID{0000-0002-4861-1412}}
\authorrunning{Schlegel et al.}
%
\institute{LMU Munich, Germany\\
\email{\{schlegel,seidl\}@dbs.ifi.lmu.de}\\
\and
Munich Center for Machine Learning (MCML), Germany
}
\maketitle              
\begin{abstract}
Counterfactual explanations emerge as a powerful approach in explainable AI, providing what-if scenarios that reveal how minimal changes to an input time series can alter the model's prediction. 
This work presents a survey of recent algorithms for counterfactual explanations for time series classification. 
We review state-of-the-art methods, spanning instance-based nearest-neighbor techniques, pattern-driven algorithms, gradient-based optimization, and generative models. 
For each, we discuss the underlying methodology, the models and classifiers they target, and the datasets on which they are evaluated. 
We highlight unique challenges in generating counterfactuals for temporal data, such as maintaining temporal coherence, plausibility, and actionable interpretability, which distinguish the temporal from tabular or image domains.
We analyze the strengths and limitations of existing approaches and compare their effectiveness along key dimensions (validity, proximity, sparsity, plausibility, etc.). 
In addition, we implemented an open-source implementation library, Counterfactual Explanations for Time Series (CFTS), as a reference framework that includes many algorithms and evaluation metrics. 
We discuss this library's contributions in standardizing evaluation and enabling practical adoption of explainable time series techniques. 
Finally, based on the literature and identified gaps, we propose future research directions, including improved user-centered design, integration of domain knowledge, and counterfactuals for time series forecasting.

\keywords{Time Series Classification \and Explainable AI \and Interpretability \and  What-If Explanations \and Counterfactuals \and Contrastive Explanations}
\end{abstract}
\section{Introduction}
Machine learning models for time series classification are increasingly applied in critical domains such as healthcare (e.g., ECG rhythm diagnosis), finance (e.g., anomaly detection in stock prices), and industrial applications (e.g., predictive maintenance)~\cite{theissler_explainable_2022}. 
In these settings, understanding why a model made a particular prediction is as important as the prediction's accuracy~\cite{rudin_stop_2019}. 
Counterfactual explanations (CFEs) have gained prominence as an explainable AI (XAI) technique to answer what-if questions: they identify how an input instance could be minimally modified to achieve a different outcome~\cite{guidotti_counterfactual_2022}. 
For example, for a patient's heart rate time series classified as "at risk", a counterfactual explanation might demonstrate how a slight reduction of a certain spike in the series would have led the model to predict a normal outcome~\cite{theissler_explainable_2022}. 
Counterfactuals are appealing because they offer actionable insight~\cite{keane_good_2020}. 
They suggest concrete changes to the input that would alter the model's decision, potentially guiding interventions or insights in learned patterns~\cite{guidotti_counterfactual_2022}.

While counterfactual explanation methods have been well-studied for tabular data~\cite{pawelczyk_carla_2021}, applying them to time series data is more challenging~\cite{delaney_instance_2021}. 
Time series classification (TSC) involves sequential, often correlated data points, so any perturbation must consider temporal dependencies to remain realistic~\cite{delaney_instance_2021}. 
Changing a value at one time step can influence future time steps, making simple counterfactual generation algorithms unplausible for users (e.g., a counterfactual ECG with an abrupt peak)~\cite{delser_generating_2024}. 
Additionally, possible real-world use of time series CFEs often implies algorithmic recourse: guiding users on how to change behavior to achieve a different outcome, such as suggesting lifestyle changes to prevent unfavourable events~\cite{chukwu_counterfactual_2025,pawelczyk_carla_2021}. 
This possible dual use of CFEs, as both explanations and recommendations, imposes extra requirements like plausibility and user-specific feasibility that go beyond flipping the model's outcome.

Over the past few years, a growing number of methods have been proposed to generate counterfactual explanations for time series classifiers~\cite{theissler_explainable_2022}. 
These methods vary in their approach and assumptions. 
Some techniques adapt distance-based optimization originally developed for tabular data, for instance, extending the loss from Wachter et al.~\cite{wachter_counterfactual_2017} to time series by incorporating channel-wise distances~\cite{ates_counterfactual_2021}.
Other approaches leverage the structure of time series by using case-based reasoning, e.g., finding a close example from a different class and exchanging its salient subsequence to serve as a counterfactual, an idea introduced as Native Guide~\cite{delaney_instance_2021}.
There are pattern-driven methods that exploit shapelets, motifs, or discords, repetitive or anomalous subsequences, to guide counterfactual generation. 
Meanwhile, researchers have explored evolutionary algorithms that search the space of possible time series perturbations using genetic operations, as well as deep learning approaches such as training generative models to produce plausible counterfactuals. 
This diversity of strategies reflects the complexity of the task and the lack of a one-size-fits-all solution.

In this work, we review the current state of algorithms for counterfactual explanations for time series classification. 
We provide a structured review of prominent methods, emphasizing how they generate counterfactuals, which models or scenarios they target, and what datasets and metrics are used for evaluation. 
We highlight the unique challenges that time series data bring to counterfactual generation, challenges of temporal coherence, plausibility, and user actionability that are less prominent in other domains. 
We then compare the strengths and limitations of different approaches in addressing these challenges. 
To do so, we implemented an open-source library, CFTS (Counterfactual Explanation Algorithms for Time Series Models)\footnote{The package is available on Github under\newline \href{https://github.com/visual-xai-for-time-series/counterfactual-explanations-for-time-series}{github.com/visual-xai-for-time-series/counterfactual-explanations-for-time-series}}, which implements many state-of-the-art algorithms under a common framework. 
By examining results and visualizations from this library, we illustrate differences between methods. 
Finally, we outline promising directions for future research, including integrating causal reasoning and improving scalability. 
Through this comprehensive review, we aim to highlight the progress made so far and identify the gaps that remain open for creating effective, human-centered counterfactual explanations for time series.

\section{Related Work and Background}

\textbf{Counterfactual Explanations --} 
The concept of counterfactual explanations in machine learning was popularized by Wachter et al.~~\cite{wachter_counterfactual_2017} as a way to explain individual predictions without revealing model internals. 
A counterfactual is typically defined as a perturbed input $\mathbf{x}'$ that yields a different output (prediction) $y'$ from the original input $\mathbf{x}$ with output $y$, while $\mathbf{x}'$ is as "close" to $\mathbf{x}$ as possible. 
Formally, one can frame finding a counterfactual as an optimization problem minimizing a weighted sum of (a) a validity loss that encourages the model’s predicted class to switch to the desired target, and (b) a proximity term that measures distance between $\mathbf{x}$ and $\mathbf{x}'$~~\cite{wachter_counterfactual_2017}. 
In practice, various distance metrics ($L_1$, $L_2$, ...) or domain-specific measures can be used to quantify the size of the change. 
Additionally, constraints or regularizers are often applied to promote sparsity (changing as few features or time points as possible), and feasibility (avoiding changes that violate domain knowledge or physical possibility)~~\cite{delser_generating_2024}.

\textbf{Time Series Classification --}
Explaining time series models poses distinct challenges because time series data inherently have a temporal order. 
A time series classifier $f(\mathbf{x})$ is often implemented as a deep neural network (CNN or RNN), trained on sequences. 
Time series often arise from processes in which only certain patterns or contiguous segments are semantically meaningful (e.g., an arrhythmia segment in an ECG)~\cite{theissler_explainable_2022}. 
Unlike tabular data, where any feature can often be changed independently, in time series, an intervention at one time point may necessitate correlated changes in other time points to maintain a realistic pattern~\cite{schlegel_interactive_2024}. 
Moreover, many time series datasets have multivariate (multi-channel) inputs, which adds complexity; counterfactuals might involve modifying one or more channels, raising questions about whether to treat channels independently or consider their joint dynamics~\cite{theissler_explainable_2022}.
Common benchmark datasets used in literature come from the UCR/UEA Time Series Archive and similar collections~\cite{dau_ucr_2019,bagnall_uea_2018}.
These include a wide range of domains (sensor readings, medical signals, synthetic control series, etc.), for example: ECG5000 (heart electrical signals)~\cite{dau_ucr_2019}, FordA (automotive sensor)~\cite{dau_ucr_2019}, Epilepsy (multichannel EEG)~\cite{bagnall_uea_2018}, and Spoken Arabic Digit dataset (multiple-channel audio traces)~\cite{bagnall_uea_2018}. 

\textbf{Existing Surveys and Position Papers --} 
Counterfactual explanations in general (not specific to time series) have been surveyed by multiple authors~\cite{guidotti_counterfactual_2022,keane_good_2020,pawelczyk_carla_2021}.
These works review methods across domains and discuss properties like actionability, sparsity, and evaluation metrics. 
However, until recently, the time series domain received relatively little attention in these surveys. 

A very recent position paper by Chukwu et al.~~\cite{chukwu_counterfactual_2025} argues that counterfactual explanations for time series must be human-centered and temporally coherent, noting that many current techniques struggle with those aspects. 
Our work complements such perspectives by providing a deep dive specifically into time series counterfactual methods, summarizing the technical progress to date, and pinpointing how well they meet the desired criteria.

In the following sections, we organize the literature by methodological categories and highlight representative techniques. 
\autoref{tab:cfts_methods} gives a high-level overview of prominent counterfactual explanation methods for time series classification, including the mechanism they use and whether they apply to univariate (U) or multivariate (M) data. 
We then discuss each category in detail.

\begin{table*}[h!]
\centering
\scriptsize
\resizebox{\textwidth}{!}{
\begin{tabular}{p{3.2cm} p{1.2cm} p{1.2cm} p{3.0cm} p{5.0cm}}
\hline
\textbf{Method} &
\textbf{Year} &
\textbf{Data} &
\textbf{Category} &
\textbf{Core Idea} \\

\hline

Wachter et al.~\cite{wachter_counterfactual_2017} &
2017 &
U/M &
Optimization-based &
Input-space loss minimization to induce class change \\

CoMTE~\cite{ates_counterfactual_2021} &
2021 &
M &
Optimization-based &
Channel-wise counterfactual optimization for multivariate time series \\

TS-Tweaking~\cite{karlsson_locally_2020} &
2020 &
U &
Optimization-based &
Greedy optimization that tweaks shapelet-aligned segments \\

TSCF &
2024 &
U/M &
Optimization-based &
Custom input-space counterfactual optimization framework \\

MOC~\cite{dandl_multi_2020} &
2020 &
U/M &
Evolutionary &
Multi-objective evolutionary counterfactual search \\

TSEvo~\cite{hoellig_tsevo_2022} &
2022 &
U/M &
Evolutionary &
Evolutionary counterfactual explanations tailored to time series \\

Sub-SpaCE~\cite{refoyo_sub_2024} &
2023 &
U &
Evolutionary &
Sparse evolutionary search over contiguous subsequences \\

Multi-SpaCE~\cite{refoyo_multi_2024} &
2024 &
M &
Evolutionary &
Multi-objective subsequence-based CFs for multivariate time series \\

Native Guide~\cite{delaney_instance_2021} &
2021 &
U &
Instance-based &
Nearest unlike neighbor–guided subsequence replacement \\

CELS~\cite{li_cels_2023} &
2023 &
U &
Instance-based &
Saliency-guided counterfactual edits for univariate series \\

M-CELS~\cite{li_m_2024} &
2024 &
M &
Instance-based &
Multivariate extension of saliency-guided CFs \\

AB-CF~\cite{li_attention_2023} &
2023 &
M &
Instance-based &
Attention-guided counterfactual explanation framework \\

Latent-CF~\cite{balasubramanian_latent_2020} &
2020 &
U/M &
Latent space &
Latent-space counterfactual baseline using autoencoders \\

CGM~\cite{vanlooveren_conditional_2021} &
2021 &
U/M &
Latent space &
Conditional generative modeling for counterfactual explanations \\

LASTS~\cite{guidotti_explaining_2020} &
2020 &
U/M &
Latent space &
Latent surrogate explanations for time series classifiers \\

GLACIER~\cite{wang_glacier_2024} &
2024 &
U/M &
Latent space &
Locally constrained, realism-aware latent counterfactuals \\

CounTS~\cite{yan_self_2023} &
2023 &
U/M &
Latent space &
Structured generative CFs with interpretable latent variables \\

SG-CF~\cite{li_sg_2022} &
2022 &
U/M &
Segment-based &
Shapelet-guided subsequence counterfactual explanations \\

SETS~\cite{bahri_shapelet_2022} &
2022 &
U/M &
Segment-based &
Efficient shapelet-based counterfactual generation \\

DisCOX~\cite{bahri_discord_2024} &
2024 &
U/M &
Segment-based &
Discord (anomalous segment) replacement for CFs \\

CFWoT~\cite{sun_counterfactual_2024} &
2024 &
M &
Segment-based &
Subsequence-based CFs without access to training data \\

TS-CEM~\cite{labaien_contrastive_2020} &
2020 &
U/M &
Segment-based &
Time-series adaptation of CEM with contrastive temporal segments \\

MG-CF~\cite{li_motif_2022} &
2022 &
U/M &
Hybrid &
Motif-guided counterfactuals combining segment and instance ideas \\

SPARCE~\cite{lang_generating_2023} &
2022 &
M &
Hybrid &
Structured sparsity for actionable counterfactual recourse \\

TeRCE~\cite{bahri_temporal_2022} &
2022 &
M &
Hybrid &
Symbolic temporal-rule-based counterfactual explanations \\

Time-CF~\cite{huang_shapelet_2024} &
2024 &
U/M &
Hybrid &
GAN-based counterfactuals guided by temporal shapelets \\

\end{tabular}
}
\vspace{0.5em}
\caption{Overview of counterfactual explanation methods for time series classification, including publication year and supported data type (U: univariate, M: multivariate).}
\label{tab:cfts_methods}
\end{table*}

\section{Collected Methods for Counterfactual Explanations in Time Series Classification and Others}
\label{sec:methods}

To provide a structured overview of existing approaches, we organize the literature into six methodological categories based on how counterfactuals are generated and the structural constraints imposed on the time series. 
This taxonomy emphasizes the underlying mechanism rather than specific applications and enables principled comparison across methods. 
The categories are: 
\textbf{
(i) optimization-based methods, 
(ii) evolutionary methods, 
(iii) instance-based methods, 
(iv) latent space methods, 
(v) segment-based methods, and 
(vi) hybrid methods.
}
\autoref{tab:cfts_methods} summarizes implemented works in CFTS within each category with basic concepts.

\textbf{Methodology --}
Our overview builds upon the work of Theissler et al.~\cite{theissler_explainable_2022} and Chukwu et al.~\cite{chukwu_counterfactual_2025}, which provide comprehensive overviews of counterfactual explanation methods in machine learning. 
Using these works as a conceptual and structural foundation, we extended the literature search to identify more recent, domain-specific studies. 
To this end, we conducted a systematic search on Google Scholar using keywords related to counterfactual explanations and time series data.
Additionally, we examined the citation networks of one of the earliest counterfactual works for time series to uncover subsequent research and extensions. 
This combination of targeted keyword search and citation tracing ensured broad coverage of both seminal and emerging contributions in the field.

\subsection{Optimization-Based Methods}

Optimization-based methods formulate counterfactual generation as a direct optimization problem in the input space.
Given an input time series and a trained classifier, these approaches search for a modified time series that minimizes a combination of 
(i) a validity objective, enforcing the model to predict a desired target class, and 
(ii) a proximity objective, measuring the distance to the original instance.
These approaches usually operate in the input space and either assume access to the classifier's gradients or rely on score-based approximations, where a surrogate density model is estimated to approximate the gradient and used to guide the search for valid yet minimally perturbed counterfactuals.

The formulation introduced by Wachter et al.~\cite{wachter_counterfactual_2017} serves as the conceptual foundation of this family. 
Although originally proposed for tabular data, it has been widely adapted to time series by defining suitable distance measures and update rules. 
When applied to temporal data, Wachter-style optimization allows fine-grained pointwise adjustments, enabling counterfactuals with very small distances. 
However, this often leads to temporally diffuse changes, where small perturbations are spread across many time steps, limiting interpretability.

CoMTE~\cite{ates_counterfactual_2021} extends optimization-based counterfactual generation to multivariate time series by explicitly performing channel-wise optimization. 
Instead of treating all dimensions jointly, CoMTE evaluates which subset of channels must be modified to achieve a class change, thereby improving interpretability in settings where channels correspond to distinct physical sensors. 
This design helps identify which variables matter, but it operates at a relatively coarse temporal granularity and may miss fine-grained temporal patterns within channels.

TS-Tweaking~\cite{karlsson_locally_2020} formulates explainable time series tweaking as an optimization problem that seeks minimal changes to an input series so that a random shapelet forest (target model) flips its prediction. 
The local variant greedily edits shapelet-aligned segments for a single instance, while the global variant derives reusable transformation rules by clustering time series and learning common tweaks that generalize across many instances.

TSCF, a custom, optimization-based framework, follows a similar approach but offers greater flexibility in defining temporal constraints and distance measures. 
By incorporating domain-specific regularization or smoothness penalties, TSCF can partially address the temporal incoherence observed in naïve input-space optimization approaches.

\textbf{Advantages --}
Optimization-based methods are computationally efficient, relatively simple to implement, and good at producing low-distance counterfactuals. 
They can be applied model-agnostically when gradients are approximated.

\textbf{Limitations --}
Without strong temporal constraints, these methods often produce diffuse, hard-to-interpret changes and struggle to balance proximity with temporal sparsity, particularly for long time series.

\subsection{Evolutionary Methods}
Evolutionary methods frame counterfactual generation as a multi-objective optimization problem solved using population-based search algorithms. 
Rather than optimizing a single weighted objective, these methods seek a set of Pareto-optimal counterfactuals that balance competing criteria such as validity, proximity, sparsity, and temporal compactness.

The MOC framework by Dandl et al.~\cite{dandl_multi_2020} introduced multi-objective counterfactual reasoning, establishing principles that later inspired time series–specific adaptations. 
TSEvo~\cite{hoellig_tsevo_2022} extends this framework to time series classification by defining mutation and crossover operators tailored to temporal data. 
Candidate time series are iteratively evolved, with fitness functions evaluating multiple explanation objectives to find fitting counterfactuals.

To improve interpretability, Sub-SpaCE~\cite{refoyo_sub_2024} restricts evolutionary edits to a small number of contiguous subsequences. 
This encourages the algorithm to find explanations that modify localized temporal regions rather than individual points scattered across time. 
Multi-SpaCE~\cite{refoyo_multi_2024} further extends this idea to multivariate time series, allowing simultaneous subsequence edits across multiple channels while preserving temporal alignment.

Evolutionary methods produce diverse counterfactual explanations, offering users multiple plausible alternatives. 
This diversity is particularly valuable in decision-support scenarios. 
However, population-based search incurs a high computational cost, as each generation requires numerous model evaluations.

\textbf{Advantages --}
Evolutionary methods excel at producing sparse, localized, and diverse counterfactuals, and are fully model-agnostic.

\textbf{Limitations --}
They are computationally expensive and may not scale well to long, high-dimensional, or real-time settings.

\subsection{Instance-Based Methods}

Instance-based methods generate counterfactual explanations by incorporating real examples from the dataset. 
The central assumption is that plausible counterfactuals resemble existing instances from the target class.

Native Guide Counterfactuals (NG-CF)~\cite{delaney_instance_2021} identify a nearest unlike neighbor from the target class and adapt salient subsequences to form a counterfactual. 
Importance scores or attribution methods are often used to determine which parts of the time series should be replaced. 
This results in temporally coherent counterfactuals composed of realistic patterns.

Building on this idea, CELS~\cite{li_cels_2023} introduces saliency-guided counterfactual generation, where learned importance maps restrict modifications to influential time steps. 
M-CELS~\cite{li_m_2024} extends this approach to multivariate time series, while AB-CF~\cite{li_attention_2023} leverages attention mechanisms within neural classifiers to guide counterfactual edits across channels and time.

Instance-based approaches naturally preserve plausibility and are highly interpretable, as the changes correspond to known examples. 
However, their success depends on the availability of suitable target-class neighbors and may degrade when dataset coverage is limited.

\textbf{Advantages --}
Because counterfactuals are constructed from real observed instances, instance-based methods tend to align well with domain expectations and are often easier for practitioners to trust and validate.

\textbf{Limitations --}
Their reliance on nearest neighbors can also introduce bias toward frequently observed patterns, potentially overlooking rare but valid counterfactual alternatives.

\subsection{Latent Space Methods}

Latent space methods operate on learned representations of time series rather than directly manipulating raw inputs. 
These representations, typically learned via autoencoders or generative models, capture global temporal structure and statistical dependencies.

Latent-CF~\cite{balasubramanian_latent_2020} provides a simple baseline by generating counterfactuals in an autoencoder’s latent space and decoding them back to the input domain. 
CGM~\cite{vanlooveren_conditional_2021} extends this idea using conditional generative models, allowing counterfactuals to be sampled from a learned class-conditional distribution. 
LASTS~\cite{guidotti_explaining_2020} introduces surrogate-based explanations, approximating a black-box classifier with an interpretable latent model.

More recent work, such as GLACIER~\cite{wang_glacier_2024}, introduces locally constrained latent optimization that explicitly enforces smoothness, plausibility, and neighborhood consistency. 
CounTS~\cite{yan_self_2023} embeds counterfactual reasoning into a structured generative model, separating mutable and immutable latent factors and enabling self-interpretable explanations.

Latent space methods are particularly effective at maintaining plausibility and global temporal coherence but rely heavily on the quality and interpretability of the learned representations.

\textbf{Advantages --}
By operating on compressed representations, latent space methods can capture long-range temporal dependencies that are difficult to enforce explicitly in input-space approaches.

\textbf{Limitations --}
The abstract nature of latent variables may obscure the meaning of the suggested changes, complicating their use for actionable recourse.

\subsection{Segment-Based Methods}

Segment-based methods restrict counterfactual modifications to contiguous subsequences, aligning explanations with human intuition about temporal events. 
These subsequences may correspond to discriminative shapelets, recurring motifs, or anomalous segments.

SG-CF~\cite{li_sg_2022} and SETS~\cite{bahri_shapelet_2022} rely on shapelet representations to identify informative subsequences and modify them to induce class changes.
DisCOX~\cite{bahri_discord_2024} focuses on replacing discord (anomalous) segments that disproportionately influence the classifier. 
CFWoT~\cite{sun_counterfactual_2024} enables subsequence-level edits even in the absence of training data, making it suitable for data-limited scenarios.

TS-CEM~\cite{labaien_contrastive_2020} extends CEM’s~\cite {dhurandhar_explanations_2018} contrastive explanation framework to deep time series classifiers by optimizing over discriminative temporal segments rather than pixels or tabular features. 
For a given series and predicted class, TS-CEM searches for contiguous subsequences whose removal or alteration would flip the prediction (pertinent positives/negatives), yielding contrastive “why this vs.\ that” explanations grounded in specific time intervals.

By focusing on segments rather than individual points, these methods yield compact and interpretable counterfactuals. 
However, their success depends on robust mechanisms for pattern discovery.

\textbf{Advantages --}
Restricting edits to contiguous segments allows these methods to express explanations in terms of patterns rather than isolated values.

\textbf{Limitations --}
If the decision boundary depends on subtle global trends rather than localized patterns, segment-based constraints may fail to identify valid counterfactuals.

\subsection{Hybrid Methods}

Hybrid methods combine elements from multiple methodological families to leverage complementary strengths. 
These approaches often integrate generative modeling, pattern-based constraints, symbolic reasoning, or sparsity objectives.

MG-CF~\cite{li_motif_2022} combines motif-based reasoning with instance-based ideas, while SPARCE~\cite{lang_generating_2023} introduces structured sparsity constraints to generate actionable recourse. 
TeRCE~\cite{bahri_temporal_2022} uses symbolic temporal rules to express counterfactuals at a high level, and Time-CF~\cite{huang_shapelet_2024} integrates generative adversarial modeling with shapelet guidance to produce realistic yet interpretable counterfactuals.

Hybrid methods are highly flexible and expressive, making them suitable for complex and high-stakes applications, though they often introduce additional modeling complexity.

\textbf{Advantages --}
By combining complementary mechanisms, hybrid methods can flexibly adapt to different datasets and explanation objectives without being tied to a single inductive bias.

\textbf{Limitations --}
This flexibility can make hybrid approaches harder to analyze theoretically and more sensitive to design choices and hyperparameter settings.

\subsection{Summary}

This extended taxonomy highlights the diversity of counterfactual explanation methods for time series classification and clarifies their respective trade-offs. 
Optimization-based methods favor proximity; evolutionary methods emphasize diversity and sparsity; instance- and segment-based methods excel in interpretability; latent space methods enhance plausibility; and hybrid approaches aim to unify these strengths. 
This structured perspective enables systematic evaluation and provides a solid foundation for future research.

\section{Challenges Unique to Time Series Counterfactuals}

Designing algorithms for counterfactual explanations for time series has unique challenges that researchers must address to make these explanations useful and trustworthy. 
We highlight several key challenges and considerations:

\textbf{Temporal Coherence --} 
Any counterfactual modification must respect the temporal nature of time series. 
Changing a single time step in isolation can break temporal patterns or introduce inconsistencies (e.g., a sudden jump that violates physical laws). 
Counterfactuals should ideally modify contiguous segments to keep the overall sequence smooth. 
Temporal coherence also means accounting for lead-lag relationships. 
This means that if a cause precedes an effect in the original series, a meaningful counterfactual likely needs to reflect a plausible adjustment to the following time steps, not just an arbitrary perturbation.

\textbf{Plausibility --} 
A counterfactual time series should "look like" a time series that could have been observed in the real world (preferably of the target class). 
This is particularly crucial in domains like medicine: a generated ECG that contains physiologically impossible waveform shapes would not be credible to a doctor. 
Plausibility can involve simple value constraints (e.g., no negative values for features that can’t be negative) and more complex statistical properties (preserving autocorrelation structure, frequency content, or known pattern signatures of the phenomenon)~\cite{delser_generating_2024}. 
Many methods address this implicitly by drawing on real data (instance-based, segment-based), or explicitly via constraints and generative models. 
Nonetheless, ensuring plausibility remains challenging, especially for multivariate series where inter-variable correlations must be maintained~\cite{delser_generating_2024}.
Measures such as domain constraint violations and statistical similarity have been proposed to quantify plausibility, but not all studies evaluate them, risking counterfactuals that optimize the loss but would never occur in reality.

\textbf{Actionability and User Constraints --}
In time series contexts, we often interpret counterfactuals as recommendations for action, e.g., "if the patient had walked 1000 extra steps each day, their glucose level time series would fall into a normal pattern."~\cite{chukwu_counterfactual_2025}
For such recommendations to be actionable, we must avoid suggesting changes to immutable aspects of the series (like past events or static attributes). 
We should also consider whether the suggested change can be feasibly carried out by the user or the system. 
For instance, reducing a manufacturing sensor’s vibration at a specific cycle might require a specific intervention, which has a cost. 
The feasibility of counterfactuals is a challenge: ensuring that each change corresponds to a real intervention and that the magnitudes are reasonable.
This ties into causal considerations. 
One should generate only those counterfactuals that correspond to interventions in the domain’s causal model (changing an effect without changing its cause would be infeasible). 
Current methods rarely integrate user-specific constraints automatically, leaving this an important gap in making CFEs truly useful in real-world applications.

\textbf{Sparsity vs. Context --}
A longstanding principle in counterfactual explanation is that changes should be sparse. 
However, in time series, sparsity needs to be balanced with temporal context. 
Changing a single time point might be minimal, but it could also be meaningless (or subtle in a sensor signal). 
Many time series CF methods thus aim for segment-level sparsity, altering a small number of contiguous segments or a key pattern. 
This yields what some metrics call compactness, measuring whether changes occur in a small part of the time series. 
Achieving high compactness is challenging with continuous optimization, as it often distributes small changes widely; techniques such as pattern-based constraints have been shown to achieve greater compactness. 
The challenge is to identify the right granularity for "sparsity", point-level for some domains vs. pattern-level for others, and to enforce it without sacrificing validity.

\textbf{Evaluation Metrics for Time Series CF --} 
Evaluating counterfactual quality in time series requires going beyond the traditional validity, proximity, and sparsity.
There is a temporal blind spot in many evaluation protocols: metrics borrowed from tabular settings (such as $L_1/L_2$ distances or the number of features changed) do not capture how changes are distributed over time. 
For example, two counterfactuals might have the same $L_2$ distance from the original, but one might make a one-time shift of a segment, while the other adds high-frequency noise throughout; clearly, the former is more interpretable.
Metrics like Dynamic Time Warping (DTW) distance can measure similarity, allowing for slight time shifts, which might be relevant if the timing of patterns can vary. 
Other specialized metrics include temporal inconsistency (whether the time ordering of events is distorted) and shape-based distance. 
Additionally, plausibility metrics, as noted above, are critical to measure. 
Some benchmarking efforts propose multi-faceted evaluations covering validity, proximity, sparsity, plausibility, diversity, stability, etc ~\cite{delser_generating_2024,delaney_counterfactual_2023,keane_if_2021} 
A challenge is that optimizing for all is impossible, so understanding trade-offs is key. 
The lack of universal metrics for temporal plausibility makes comparing methods across papers difficult.
One method might claim a better $L_2$ distance, another better sparsity. 
To truly move forward, the field needs evaluation frameworks that reflect the temporal dimension.

\textbf{User Understanding and Trust --}
Beyond technical metrics, a core challenge is how humans perceive and trust time series counterfactual explanations. 
A user (be it a doctor, engineer, or customer) might be presented with an alternative time series and an assertion that "this would change the prediction." 
If the counterfactual is too complex or unrealistic, the user might disregard it or, worse, lose trust in the AI system. 
There is early recognition that user studies are needed to evaluate if, say, clinicians find these explanations useful in practice~\cite{chukwu_counterfactual_2025}. 
Time series, with their often incomprehensible lines, are not as immediately interpretable as images; thus, conveying the content of a counterfactual (via visual highlights of changed regions or descriptions) is part of the challenge~\cite{schlegel_visual_2023}. 
So far, most research has focused on algorithmic aspects rather than human factors. 
Ensuring that time series CFEs are human-centered (aligned with user mental models and decision processes) remains an open challenge.

In summary, counterfactual explanations for time series must overcome issues of temporal logic, plausibility, actionable insight, and evaluation complexity. 
Many of these challenges are active areas of research. 
In the next section, we analyze how the current methods fare with respect to these issues, and where improvements are still needed.

\section{Comparative Analysis of Existing Methods}

Having reviewed the landscape of counterfactual explanation methods for time series classification, we now analyze their strengths and limitations in light of the challenges discussed. 
While all surveyed methods aim to produce valid counterfactuals, they differ substantially in how they balance competing properties such as proximity, sparsity, temporal coherence, plausibility, and computational efficiency. 
\autoref{tab:tscf_comparison} provides a qualitative comparison of major methodological families across criteria, including validity, temporal coherence, sparsity, plausibility, and computational cost. 
This comparison is conceptual in nature and is based on reported empirical results in the literature as well as our synthesized understanding of these approaches.

\begin{table}[t]
\centering
\scriptsize
\resizebox{\columnwidth}{!}{
\begin{tabular}{p{3.4cm} p{1.4cm} p{1.8cm} p{1.4cm} p{1.8cm} p{2.0cm}}
\hline
\textbf{Method Family} &
\textbf{Validity} &
\textbf{Temporal Coherence} &
\textbf{Sparsity} &
\textbf{Realism} &
\textbf{Computational Cost} \\

\hline

Optimization-based &
High &
Low--Medium &
Low &
Low--Medium &
Low \\

Evolutionary &
High &
High &
High &
Medium--High &
High \\

Instance-based &
Medium--High &
High &
Medium--High &
High &
Medium \\

Latent space &
High &
High &
Medium &
High &
Medium--High \\

Segment-based &
Medium--High &
Very High &
High &
High &
Medium \\

Hybrid &
High &
High &
High &
High &
Medium--High \\

\end{tabular}
}
\vspace{0.5em}
\caption{Qualitative comparison of counterfactual explanation method families for time series classification.
Assessments are based on reported results in the literature and conceptual analysis rather than absolute quantitative benchmarks.}
\label{tab:tscf_comparison}
\end{table}

\textbf{Validity as a Baseline Requirement --}
First, most methods prioritize validity; a counterfactual that fails to flip the model’s prediction is inherently uninformative. 
Most approaches explicitly encode validity either as a hard constraint or as a dominant optimization objective, and many report near-100\% success rates on benchmark datasets when a counterfactual can be found. 
Consequently, validity alone is rarely the differentiating factor between methods. 
Instead, practical differences emerge primarily in the remaining explanation properties.

\textbf{Proximity vs. Sparsity Trade-offs --}
A key distinction across methods lies in the trade-off between proximity and sparsity. 
Optimization-based approaches, including Wachter-style methods~\cite{wachter_counterfactual_2017} and latent-space variants such as LatentCF~\cite{balasubramanian_latent_2020}, tend to excel in proximity by finely adjusting values across the entire sequence. 
This often leads to very low $L_2$ or DTW distances. 
However, these gains typically come at the cost of sparsity: many time steps are slightly modified, resulting in diffuse changes that may be difficult to interpret.

In contrast, evolutionary, pattern-based, and segment-based methods are explicitly designed to promote sparsity and temporal compactness. 
Techniques such as Multi-SpaCE~\cite{refoyo_multi_2024}, SG-CF~\cite{li_sg_2022}, or motif-based approaches intentionally restrict modifications to one or a few contiguous regions, yielding explanations that are visually and semantically localized. 
Our empirical studies illustrate this contrast: evolutionary approaches alter as little as 10\% of a sequence, whereas unconstrained optimization methods may modify over 70\% of time points, albeit by very small magnitudes. 
These observations suggest that a multi-objective perspective is essential—users may prefer a slightly larger, localized change over a numerically smaller, globally dispersed modification. 

\textbf{Temporal Plausibility --}
Temporal plausibility represents another major differentiator. 
Instance-based and pattern-based approaches, such as NG-CF~\cite{delaney_instance_2021}, motif-guided counterfactuals, and discord replacement methods, inherently produce realistic changes because they rely on real subsequences. 
Their outputs typically preserve the global structure and reduce unnatural noise.

By contrast, unconstrained gradient-based methods can introduce subtle oscillations or artifacts across the entire sequence. 
While these artifacts may be mathematically optimal, they often fail qualitative plausibility checks. 
Recent methods such as GLACIER~\cite{wang_glacier_2024} and Time-CF~\cite{huang_shapelet_2024} address this limitation by introducing latent constraints or generative models, respectively, resulting in markedly more realistic counterfactuals as confirmed by both quantitative metrics and human evaluation.

Evolutionary methods sit between these extremes: while they do not inherently guarantee plausibility, it can be enforced through fitness terms or crossover operators that recombine real data segments, as in TSEvo~\cite{hoellig_tsevo_2022}. 
A notable limitation across nearly all methods, however, is the limited treatment of long-range temporal dependencies such as seasonality or trends.
Few approaches explicitly ensure that counterfactuals preserve these global temporal structures, highlighting an important open research direction.

\textbf{Multivariate Coordination --}
Handling multivariate time series introduces additional complexity. 
Methods such as CoMTE~\cite{ates_counterfactual_2021} and shapelet-based approaches like SETS~\cite{bahri_shapelet_2022} explicitly reason about channels independently, often identifying a subset of variables that require modification while leaving others untouched. 
This can improve interpretability when channels correspond to distinct sensors or measurements.

In contrast, approaches such as TSEvo~\cite{hoellig_tsevo_2022} or GLACIER~\cite{wang_glacier_2024} often treat multivariate inputs jointly, which can inadvertently introduce correlated changes across channels. 
This raises subtle concerns: in many real-world systems, modifying one variable may require coordinated changes in others due to physical or causal constraints. 
Most current methods do not guarantee such coordination explicitly and instead rely on the predictive model to implicitly penalize unrealistic combinations.
Consequently, the appropriateness of channel-wise versus joint modification strategies depends strongly on the application domain.

\textbf{Computational Efficiency --}
There are significant differences in computational efficiency across methods. 
Instance-based approaches and simple pattern-replacement strategies are typically fast, often requiring only nearest-neighbor retrieval and minor adjustments, making them suitable for large-scale use. 
Gradient-based optimization methods are also efficient in most cases, especially when implemented on GPUs, although latent approaches incur additional preprocessing costs due to representation learning.

Evolutionary algorithms are consistently the most computationally expensive, with reported runtimes ranging from minutes to hours per instance, depending on sequence length and population size. 
While acceptable for offline analysis, this limits their practicality in time-sensitive settings. 
Reinforcement-learning-based approaches offer an interesting compromise: although training the policy can be expensive, inference is extremely fast once the agent is learned. 
From a systems perspective, this distinction becomes critical when explanations are required at scale. 
The open-source library discussed later in this paper explicitly supports runtime benchmarking, encouraging more transparent reporting of efficiency alongside explanation quality.

\textbf{Diversity of Explanations --}
Diversity is an often underemphasized but important property of counterfactual explanations. 
Multi-objective evolutionary approaches such as MOC~\cite{dandl_multi_2020} and TSEvo~\cite{hoellig_tsevo_2022} produce sets of diverse counterfactuals, representing different ways to alter a prediction. 
This aligns well with best practices in XAI, which caution against presenting a single explanation as the sole interpretation. 
Other methods typically return a single counterfactual unless explicitly re-run with different initializations or constraints. 
Incorporating user preferences—allowing users to choose among multiple plausible counterfactuals—remains an underexplored but promising direction.

\textbf{Illustrative Example --}
To illustrate these trade-offs concretely, consider a univariate time series from the FordA~\cite{dau_ucr_2019} dataset that is initially classified as class 1. 
A gradient-based counterfactual may achieve class 2 by slightly shifting the entire time series downward, modifying nearly every time step. 
In contrast, an instance-based method such as NG-CF~\cite{delaney_instance_2021} might replace a single oscillatory segment around time 100 with a segment drawn from a class 2 example, leaving the remainder untouched. 
An evolutionary approach may generate several alternatives: one modifying the segment at time 100, another altering a valley around time 250, each providing a distinct explanation. 
The "best" counterfactual depends on context, for example, whether a specific segment corresponds to a known physical event or fault in the system.

\textbf{Summary and Outlook --}
In summary, existing counterfactual explanation methods for time series classification exhibit complementary strengths. 
Optimization-based methods excel in proximity, instance- and segment-based methods prioritize interpretability and plausibility, evolutionary approaches offer principled multi-objective trade-offs at higher computational cost, and latent or hybrid methods seek to unify these properties. 
The field is increasingly moving toward hybrid frameworks that combine generative plausibility, temporal structure awareness, and optimization efficiency.
At the same time, the diversity of methods underscores the need for standardized benchmarks, shared evaluation protocols, and open libraries to enable systematic comparison.

\section{A Unified Library for Counterfactuals for Time Series}

To support research and practical application of counterfactual explanations for time series, we developed an open-source library, Counterfactual Explanations for Time Series (CFTS).
The library serves as a unified implementation of many of the algorithms discussed in this overview, providing a common codebase and standardized interfaces. 
Below, we describe the library's scope, implemented methods, and its contribution in the context of the existing literature.

\textbf{Implemented Algorithms --}
The CFTS library is a modular PyTorch-based framework that implements a broad range of counterfactual generation techniques for time series classification, spanning all families described in~\autoref{sec:methods}. 
The goal of the library is not to exhaustively implement every published method, but to provide representative approaches that enable systematic comparison across paradigms to showcase differences.

Currently, the library includes implementations of \textit{optimization-based methods}, \textit{evolutionary methods}, \textit{instance-based methods}, \textit{latent space methods}, \textit{segment-based methods}, and \textit{hybrid methods}. For \textit{optimization-based methods}, we provide \textbf{Wachter-style}~\cite{wachter_counterfactual_2017} gradient-based counterfactuals, a genetic-algorithm variant of Wachter-style optimization, \textbf{CoMTE}~\cite{ates_counterfactual_2021} for multivariate time series, \textbf{TS-CEM}~\cite{labaien_contrastive_2020} as a CEM-style temporal segment approach, and \textbf{TSCF} for input-space optimization with temporal smoothness and sparsity. The \textit{evolutionary methods} currently implemented are \textbf{MOC} (Multi-Objective Counterfactuals)~\cite{dandl_multi_2020}, \textbf{TSEvo}~\cite{hoellig_tsevo_2022} for evolutionary time series counterfactuals, and \textbf{Sub-SpaCE}~\cite{refoyo_sub_2024} together with \textbf{Multi-SpaCE}~\cite{refoyo_multi_2024} for sparse subsequence search in uni- and multivariate series. For \textit{instance-based methods}, the library includes \textbf{Native Guide} counterfactuals~\cite{delaney_instance_2021}, saliency-guided methods \textbf{CELS}~\cite{li_cels_2023} and \textbf{M-CELS}~\cite{li_m_2024}, and \textbf{AB-CF}~\cite{li_attention_2023} which leverages attention weights from neural classifiers. The \textit{latent space methods} comprise \textbf{Latent-CF}~\cite{balasubramanian_latent_2020} as a latent-space baseline, \textbf{CGM}~\cite{vanlooveren_conditional_2021} for class-conditional counterfactual sampling, and \textbf{GLACIER}~\cite{wang_glacier_2024} for locally constrained, plausibility-aware latent optimization. Among \textit{segment-based methods}, we implement shapelet-based counterfactuals \textbf{SETS}~\cite{bahri_shapelet_2022} and \textbf{SG-CF}~\cite{li_sg_2022}, discord-based counterfactuals \textbf{DisCOX}~\cite{bahri_discord_2024}, and \textbf{TS-Tweaking}~\cite{karlsson_locally_2020} for greedy tweaking of shapelet-aligned segments. Finally, the \textit{hybrid methods} include \textbf{Time-CF}~\cite{huang_shapelet_2024} combining generative modeling with pattern guidance, \textbf{MG-CF}~\cite{li_motif_2022} integrating motif-guided, instance-, and segment-based reasoning, \textbf{TeRCE}~\cite{bahri_temporal_2022} for symbolic temporal-rule-based counterfactuals, and \textbf{SpArCE}~\cite{lang_generating_2023} with structured sparsity for actionable recourse.

Overall, CFTS currently implements a diverse and methodologically complete set of counterfactual explanation techniques, covering all six categories introduced in this overview. 
Each algorithm is encapsulated in its own module with a shared interface, enabling consistent execution, evaluation, and visualization across methods. 
This level of coverage makes CFTS a comprehensive open-source library for counterfactual explanations in time series classification.
By consolidating these approaches under a single framework with standardized inputs and outputs, CFTS removes a major barrier to reproducibility and benchmarking that has historically limited progress in this area~\cite{bhattacharya__2024}.

\textbf{Evaluation Metrics and Tools --}
A central contribution of CFTS is its extensive evaluation framework, designed to reflect the multifaceted nature of counterfactual explanations. 
Metrics are grouped into six categories:
\begin{itemize}
    \item \textbf{Validity metrics}, ensuring that the counterfactual achieves the desired target class.
    \item \textbf{Proximity metrics}, including $L_1$, $L_2$, Dynamic Time Warping (DTW), and Fréchet distance, capturing both pointwise and shape-based similarity.
    \item \textbf{Sparsity and compactness metrics}, such as $L_0$ (number of modified points), segment count, and modified segment length.
    \item \textbf{Plausibility metrics}, measuring distributional similarity, autocorrelation preservation, and spectral consistency.
    \item \textbf{Diversity metrics}, quantifying differences among multiple counterfactuals generated for the same input.
    \item \textbf{Stability metrics}, assessing robustness to small perturbations of the input.
\end{itemize}

The inclusion of DTW and spectral measures reflects an important design choice: in time series analysis, shifting a pattern slightly in time may be semantically acceptable and should not be penalized as harshly as entirely changing its structure. 
By making such metrics readily available, CFTS encourages researchers to move beyond single-number evaluations and report further results to ensure counterfactual quality.

\begin{figure}
    \centering
    \includegraphics[height=0.95\textheight]{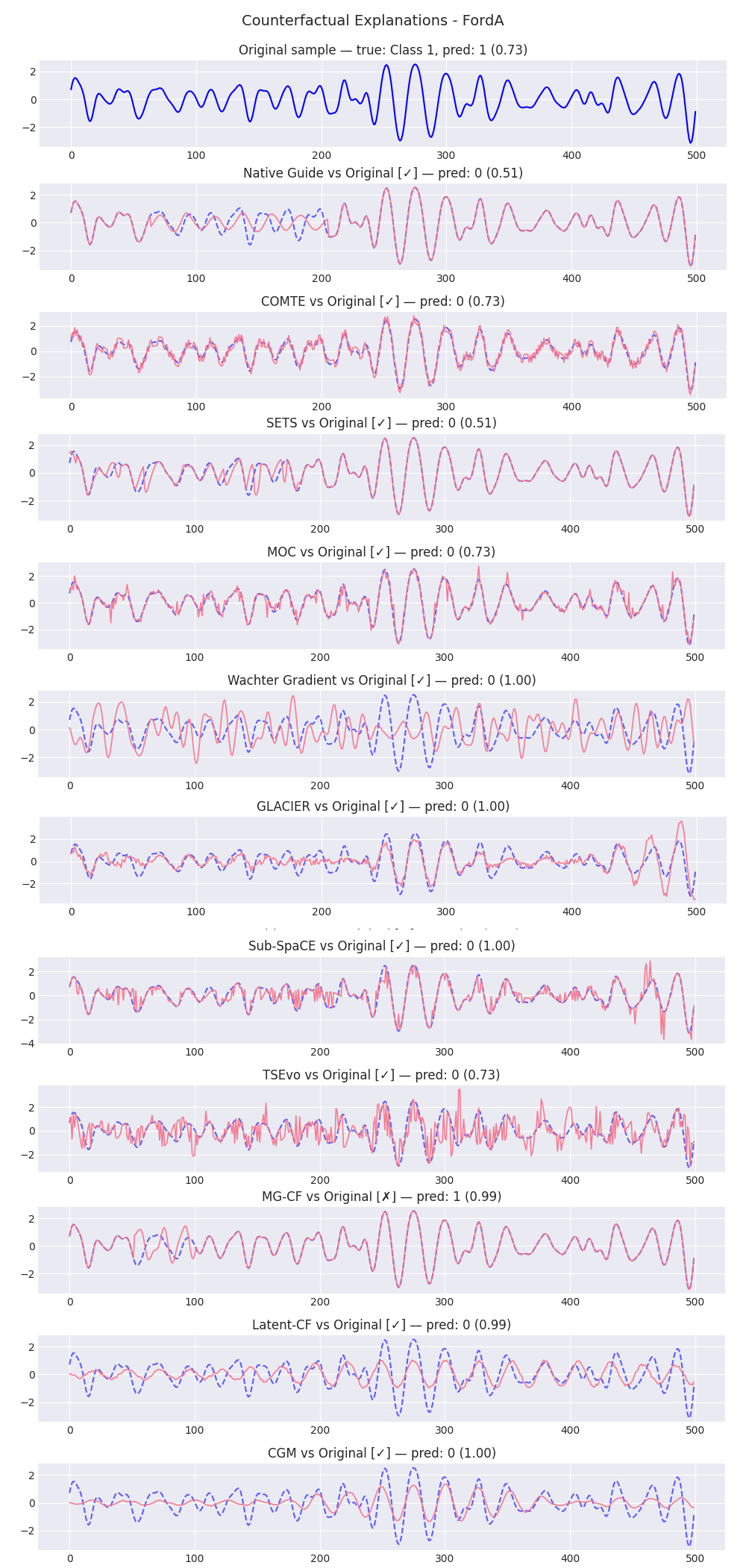}
    \caption{A FordA sample with various counterfactual methods applied to a CNN trained to classify the data between normal and anomaly classes.}
    \label{fig:cf_example}
\end{figure}

\textbf{Visualization and User Interface --}
CFTS provides built-in visualization methods to support qualitative analysis. 
These include overlays of original and counterfactual time series that highlight modified regions, as well as side-by-side comparisons across multiple methods. 
\autoref{fig:cf_example} illustrates an example for the FordA dataset, where multiple counterfactuals achieve the same class flip but differ in the segments they modify.

For multivariate data, the library supports channel-wise visualizations, enabling inspection of inter-channel dependencies. 
This is particularly important for datasets such as spoken Arabic digit signals, where counterfactuals may alter only a subset of channels while preserving realistic relationships among others. 
Visualization of evaluation metrics—such as bar charts comparing proximity, sparsity, and plausibility across algorithms—is also supported, closely mirroring comparative figures commonly found in the literature.

\textbf{Practical Usability and Extensibility --}
Beyond research, the unified design of CFTS lowers the barrier for practitioners to apply counterfactual explanations to real-world time series models. 
A user can supply a trained classifier and dataset, select a method (e.g., \texttt{wachter}, \texttt{native\_guide}), and obtain counterfactual explanations through a consistent API. 
A shared \texttt{CounterfactualEvaluator} class ensures that metrics are computed uniformly across methods.

The modular architecture also facilitates extensibility. 
New algorithms can be added by implementing a standardized interface, and the repository invites community contributions. 
While the library already supports evolutionary and optimization-based methods, reinforcement-learning-based counterfactuals are not yet included, highlighting an opportunity for future extensions.

\textbf{Context in the Literature --}
The CFTS library aligns with a broader movement toward benchmarking and standardization in explainable AI. 
Prior work has emphasized the need for systematic evaluation of counterfactual explanations, and XTSC-Bench~\cite{hoellig_xtsc_2023} has provided benchmark datasets for explainable time series classification. 
CFTS complements these efforts by providing the implementations needed to run such benchmarks in practice.

By including both foundational methods (e.g., Wachter-style counterfactuals) and recent advances (e.g., GLACIER, Multi-SpaCE, AB-CF), the library enables researchers to quantify progress across generations of methods—for example, evaluating whether newer approaches genuinely improve plausibility or sparsity and at what computational cost.



\textbf{Summary --}
In summary, CFTS represents a significant contribution to explainable AI for time series. 
It operationalizes a wide range of counterfactual explanation methods, provides a comprehensive evaluation suite, and supports both qualitative and quantitative analysis. 
By lowering barriers to reproducibility and benchmarking, the library helps bridge the gap between methodological research and real-world application. 
We recommend that future counterfactual methods for time series be evaluated within such shared frameworks to facilitate fair and transparent comparison, and we view CFTS as a foundation upon which the community can continue to build.

\section{Future Research Directions}

Our overview reveals a rapidly evolving field with many promising approaches, but also underscores open problems and opportunities. In this section, we outline several future research directions for counterfactual explanations in time series classification, based on gaps identified in the literature and emerging trends:

\textbf{User-Centered and Domain-Specific Counterfactuals --} 
A recurring theme is the need for counterfactuals that are meaningful and useful to end-users (be it doctors, engineers, or individuals subject to algorithmic decisions)~\cite{schlegel_visual_2023}. 
Future work should incorporate human-in-the-loop design~\cite{schlegel_interactive_2024}: for example, interactive tools where users can specify which features or time segments are actionable for them, and the counterfactual generation adapts accordingly. 
Domain-specific knowledge should be embedded into the counterfactual generation process, for instance, using clinical constraints in medical time series (do not suggest impossible vital signs). 
This might involve collaboration with domain experts to craft constraint rules or the use of simulators to verify the plausibility of counterfactual modifications~\cite{keane_if_2021}. 
Ultimately, conducting user studies to evaluate how well counterfactual explanations improve human decision-making and trust in each domain will guide refinements to these methods~\cite{schlegel_visual_2023}.

\textbf{Causality and Recourse in Temporal Settings --} 
Building on the initial application of causal modeling (like CounTS~\cite{yan_self_2023}), further research should integrate causal inference with time series counterfactuals. 
One direction is to use causal discovery on time series to learn which variables truly influence outcomes, and then restrict counterfactual changes to those causal drivers. 
Another is to address the non-independent and identically distributed nature of time series: a counterfactual change at time $t$ might influence the model’s prediction not just at $t$ but at future time steps in temporal models (for example, in forecasting models). 
There is a need for algorithms that provide counterfactual recourse over time, suggesting a series of actions over a timeline that will lead to a desired outcome~\cite{chukwu_counterfactual_2025}. 
This could be framed as a temporal decision-making problem or using techniques from time series causal analysis.
Additionally, questions of fairness and bias in time series models could be tackled by counterfactual methods that reveal differential treatment: e.g., does a certain type of time series pattern systematically require larger changes for one class vs. another to achieve a favorable outcome? 
Addressing causality and fairness together might involve counterfactual analysis on generative temporal causal models.

\textbf{Scalability and Real-Time Explanations --} 
As time series datasets grow in length and sampling frequency (e.g., high-resolution sensor streams), existing counterfactual methods may not scale well. 
Research should focus on computational efficiency. 
This could mean developing approximate methods that sacrifice some objectives for speed, for example, using segment-level abstraction (dividing the series into larger chunks and only optimizing those) to reduce dimensionality. 
A reinforcement learning approach could be promising here.
Once trained, it could provide near-instantaneous counterfactual generation, which is suitable for real-time systems (e.g., explaining an anomaly detection to an operator immediately so they can take action). 
Exploring other fast paradigms, like differentiable programming or heuristic search for time series, could yield methods that operate in near-linear time with series length. 
Moreover, parallel and distributed computing might be leveraged for evolutionary algorithms to handle large populations in less time. 
The field would benefit from benchmarking not just on small academic datasets, but on longer multivariate series (hundreds or thousands of time steps) to identify bottlenecks and drive optimizations.

\textbf{Integration with Other Explainability Techniques --} 
Counterfactual explanations need not exist in isolation. 
Combining them with feature attributions or example-based explanations could produce richer insights~\cite{schlegel_visual_2023}. 
For example, one might use an attribution method to first highlight which time region influenced the prediction, then generate a counterfactual focused on that region, this two-step approach could be more efficient and yield explanations that are easier to communicate ("the model focused on this spike; here’s how altering that spike changes the prediction")~\cite{schlegel_time_2021}. 
Similarly, counterfactuals could be used to validate and refine concept-based explanations: if a model claims to detect a concept (like "oscillation pattern") for classification, one could generate a counterfactual that removes that pattern to see if the prediction changes. 
Visual analytics systems that allow users to toggle between different explanation modes (global feature importance, local counterfactual examples, etc.) would also enhance interpretability~\cite{schlegel_visual_2023}. 
Some initial work in visual explainable AI for time series hints at interfaces where users can manipulate parts of a time series and see the model’s response, performing their own counterfactual experiments~\cite{schlegel_visual_2023,schlegel_interactive_2024}. 

\textbf{Robustness and Uncertainty in Counterfactuals --} 
Another research avenue is quantifying the uncertainty or robustness of counterfactual explanations. 
Time series data often contain noise, and small fluctuations might flip a borderline prediction. 
So it’s worth knowing: is the counterfactual stable under slight noise? 
Some works have started testing robustness by adding noise to the original and counterfactual to see if the prediction still flips. 
More formal approaches could use Bayesian neural networks or ensemble models to gauge confidence in the counterfactual’s effect. 
If a counterfactual only works for one specific fine-tuned change and fails if the user’s action is slightly different, it may not be a reliable recommendation. 
Developing methods to produce robust counterfactuals (ones that tolerate a range of implementation errors or natural variability) will increase their practical utility. 
This might involve optimizing for not just a single instance but a neighborhood of instances around the original, or providing a confidence interval for how much change is needed.

\textbf{Applications and Case Studies --}
We advocate for increased application-driven research that demonstrates counterfactual explainability in real-world settings, consistent with the call by Keane et al.~\cite{keane_if_2021} to evaluate explanations in use rather than in isolation. 
For instance, in healthcare, counterfactual explanations could support personalized treatment planning by illustrating how minimal changes in patient trajectories affect model predictions~\cite{wang_counterfactual_2021}; 
in finance, they may be used to analyze alternative investment or risk management scenarios; 
and in environmental science, counterfactual weather or climate trajectories could help interpret complex forecasting models. 
Importantly, deploying counterfactual explanations in applied domains is likely to surface domain-specific challenges, including regulatory constraints, ethical considerations, and the feasibility of suggested interventions, which are often overlooked in purely methodological work. 
Moreover, well-documented case studies are essential to assess whether the theoretical advantages of counterfactual explanations translate into practical utility, whether domain experts and decision-makers find them understandable, trustworthy, and actionable. 
Such empirical evidence will play a critical role in enabling the broader adoption of counterfactual explanations in real-world decision-making systems.

In summary, the future direction of research on time series counterfactual explanations should strive to make them more user-aligned, causally sound, scalable, and widely applicable. 
While existing frameworks provide a robust foundation, addressing these future research directions is essential for transitioning from purely algorithmic optimizations to counterfactual explanations that facilitate human-centric interpretation of complex temporal dynamics.

\section{Conclusion}
Counterfactual explanations for time series classification have grown from an emergent idea to a large field of study in the last few years. 
We reviewed state-of-the-art methods, ranging from nearest-neighbor case-based reasoning to sophisticated multi-objective evolutionary searches and emerging reinforcement learning approaches. 
We discussed how these methods tackle the core problem of "how can we change this time series to alter the model’s decision?" through diverse strategies, whether by swapping a key subsequence with one from another class or tweaking latent representations with gradient guidance. 
We also highlighted the distinctive challenges time series data pose: ensuring temporal coherence, maintaining plausibility, and providing actionable and sparse explanations, all while balancing multiple objectives.

Our comparative analysis showed that no single method dominates in all aspects; each comes with trade-offs. 
This underscores the importance of continued research and the value of integrated frameworks like the CFTS library, which we featured as a unifying platform for implementation and evaluation. 
By standardizing algorithms and metrics, such tools allow for clearer identification of strengths and weaknesses and for driving progress in a more organized manner.

Looking ahead, it is clear that the journey toward human-centered, explainable AI for time series is still underway. 
We envision future counterfactual methods that are even more aligned with user needs, perhaps through incorporating causal knowledge and allowing interactive exploration of “what-if” scenarios over time. 
The potential impact is significant: from helping physicians understand model-assisted diagnoses by visualizing how a patient’s vitals can affect the outcome, to enabling engineers to preemptively identify how slight changes in machine sensor patterns can prevent failures.

In summary, counterfactual explanations offer a compelling narrative form of explanation. 
They do not just say “feature A was important”, but rather “if X had been different, Y would have resulted”. 
Applying this to time series data brings unique difficulties, but also unique opportunities to guide temporal decision-making. 
We hope this overview provides a comprehensive foundation for researchers entering this area and sparks new ideas that will address current limitations. 
By combining technical innovation with a focus on real-world usability, the field of time series counterfactual explainability can mature into an indispensable component of trustworthy AI systems for temporal data.

\begin{sidewaystable*}[th!]
\centering
\scriptsize
\resizebox{\textwidth}{!}{
\begin{tabular}{p{3.2cm} p{1.1cm} p{1.1cm} p{2.4cm} p{2.6cm} p{3.6cm} p{3.8cm} p{3.8cm}}
\hline
\textbf{Method} &
\textbf{Year} &
\textbf{Data} &
\textbf{Category} &
\textbf{Model access} &
\textbf{What it changes} &
\textbf{Typical strengths} &
\textbf{Typical limitations} \\

\hline

Wachter et al. &
2017 &
U/M &
Optimization-based &
Gradients / scores &
Pointwise input values &
Low-distance counterfactuals; simple and efficient &
Diffuse temporal changes; low interpretability \\

CoMTE &
2021 &
M &
Optimization-based &
Black-box &
Whole channels or channel segments &
Identifies relevant variables in multivariate data &
Coarse temporal granularity \\

TS-Tweaking &
2020 &
U &
Optimization-based &
Black-box &
Shapelet-aligned segments and points &
Actionable tweaks with local and global transformation rules &
Restricted to random shapelet forests; NP-hard search requires heuristics \\

TSCF &
2024 &
U/M &
Optimization-based &
Black-box / gradients &
Input values with temporal constraints &
Flexible framework; customizable objectives &
Depends on hand-crafted constraints \\

MOC &
2020 &
U/M &
Evolutionary &
Black-box &
Flexible (points or segments) &
Multi-objective trade-offs; diverse solutions &
High computational cost \\

TSEvo &
2022 &
U/M &
Evolutionary &
Black-box &
Points and/or segments &
Sparse, diverse counterfactuals &
Slow for long time series \\

Sub-SpaCE &
2023 &
U &
Evolutionary &
Black-box &
Few contiguous subsequences &
High temporal compactness &
Restricted solution space \\

Multi-SpaCE &
2024 &
M &
Evolutionary &
Black-box &
Multivariate subsequences &
Sparse and interpretable multivariate CFs &
Computationally expensive \\

Native Guide (NG-CF) &
2021 &
U &
Instance-based &
Black-box &
Local subsequence replacement &
Highly realistic; data-grounded &
Fails without suitable neighbors \\

CELS &
2023 &
U &
Instance-based &
Black-box &
Salient time points &
Targeted edits via learned saliency &
Sensitive to saliency noise \\

M-CELS &
2024 &
M &
Instance-based &
Black-box &
Salient multivariate regions &
Handles multivariate importance &
Requires stable saliency learning \\

AB-CF &
2023 &
M &
Instance-based &
Model-internal (attention) &
Attention-weighted regions &
Uses model’s internal reasoning &
Model-dependent \\

Latent-CF &
2020 &
U/M &
Latent space &
Encoder–decoder &
Latent representations &
Manifold-aware counterfactuals &
Latent variables lack semantics \\

CGM &
2021 &
U/M &
Latent space &
Generative model &
Generated latent samples &
High realism and diversity &
Requires training generative models \\

LASTS &
2020 &
U/M &
Latent space &
Surrogate model &
Latent surrogate features &
Model-agnostic explanations &
Approximation error \\

GLACIER &
2024 &
U/M &
Latent space &
Encoder–decoder &
Locally constrained latent edits &
Realistic, smooth counterfactuals &
Additional optimization overhead \\

CounTS &
2023 &
U/M &
Latent space &
Model-based &
Mutable latent factors &
Feasibility-aware explanations &
Strong modeling assumptions \\

SG-CF &
2022 &
U/M &
Segment-based &
Black-box &
Shapelet subsequences &
Pattern-level interpretability &
Shapelet mining overhead \\

SETS &
2022 &
U/M &
Segment-based &
Black-box &
Shapelet-based segments &
Efficient and interpretable &
Dependent on shapelet quality \\

DisCOX &
2024 &
U/M &
Segment-based &
Black-box &
Anomalous segments &
Clear explanation for anomalies &
Limited to discord-driven decisions \\

CFWoT &
2024 &
M &
Segment-based &
Black-box &
Subsequences &
Works without training data &
Limited optimization control \\

TS-CEM &
2020 &
U/M &
Segment-based &
Black-box / gradients &
Discriminative temporal segments (PPs/PNs) &
CEM-based contrastive, segment-level explanations &
Assumes CEM-style optimization; less suited to highly diffuse decisions \\

MG-CF &
2022 &
U/M &
Hybrid &
Black-box &
Motif subsequences &
Combines realism and sparsity &
Motif extraction required \\

SPARCE &
2022 &
M &
Hybrid &
Black-box &
Structured sparse features &
Actionable recourse focus &
Complex constraint design \\

TeRCE &
2022 &
M &
Hybrid &
Rule-based &
Symbolic temporal rules &
Human-readable explanations &
Lower fidelity to complex models \\

Time-CF &
2024 &
U/M &
Hybrid &
Generator + classifier &
Generated sequences / segments &
High realism via GANs &
Training instability \\

\end{tabular}
}
\caption{Comprehensive overview of counterfactual explanation methods for time series classification and beyond. 
U/M denotes univariate (U) or multivariate (M) data. See the other table for references.}
\label{tab:cfts_methods_rotated}
\end{sidewaystable*}

\begin{credits}
\subsubsection{\ackname}
We thank the creators of the UCR/UEA Time Series Archive for enabling rigorous evaluation of these methods. 
We are also grateful to the researchers whose work we reviewed for their contributions to interpretable machine learning.
\end{credits}

%
%
%
\bibliographystyle{splncs04}
\bibliography{xai2026}
%
%
%
%
%
\end{document}